\documentclass[lettersize,journal]{IEEEtran}
\usepackage{amsmath,amsfonts}
\usepackage{algorithmic}
\usepackage{algorithm}
\usepackage{array}
\usepackage{textcomp}
\usepackage{stfloats}
\usepackage{url}
\usepackage{verbatim}
\usepackage{graphicx}
\usepackage{float}
\usepackage{bbding}
\usepackage{bm}
\usepackage{cite}
\usepackage{multirow}

\usepackage[normalem]{ulem}
\usepackage[table,xcdraw]{xcolor}
\usepackage[colorlinks,linkcolor=red,citecolor=blue]{hyperref}

\begin{document}
\useunder{\uline}{\ul}{}
\title{ Indirect-Instant Attention Optimization for \\ Crowd Counting in Dense Scenes}

\author{Suyu Han, Guodong Wang*, Donghua Liu
\thanks{This work was supported by the Natural Science Foundation of Shandong Province (No. ZR2019MF050) and the Shandong Province colleges and universities youth innovation technology plan innovation team project (No. 2020KJN011).
}
\thanks{S. Han, G. Wang and D. Liu are with the College of Computer Science and Technology, Qingdao University, Qingdao 266071, China (e-mail: mail.hansuyu@gmail.com, doctorwgd@gmail.com, jgt15169@163.com).  }
\thanks{Corresponding author: Guodong Wang }}

\maketitle
 
\begin{abstract}
One of appealing approaches to guiding learnable  parameter optimization, such as feature maps, is global attention, which enlightens network intelligence at a fraction of the cost. However, its loss calculation process still falls short:  1)We can only produce  one-dimensional ``pseudo labels'' for  attention, since the artificial threshold involved in the procedure is not robust; 2) The attention awaiting  loss calculation  is necessarily   high-dimensional, and decreasing it by convolution will inevitably introduce additional learnable parameters,  thus confusing the source of the loss.   To this end, we devise a simple but efficient Indirect-Instant Attention Optimization (IIAO) module based on    SoftMax-Attention   , which transforms high-dimensional attention map into a  one-dimensional feature map in the mathematical sense for loss calculation midway through the network, while automatically providing adaptive multi-scale fusion to  feature pyramid module. The special transformation yields relatively coarse features and, originally, the predictive fallibility of regions varies by crowd density distribution, so we tailor  the Regional Correlation Loss (RCLoss) to retrieve continuous error-prone regions and smooth  spatial information . Extensive experiments have proven that our approach surpasses previous SOTA methods in many benchmark datasets. The code and  pretrained models are publicly available in the manuscript submitted for review.
\end{abstract}


\section{Introduction}
\IEEEPARstart{D}{ense} crowd counting is defined as estimating the number of people in  an image or video clip, generally using the head as the counting unit. It is potentially of great value in important areas such as video surveillance, traffic flow control, cell counting, pest and disease prevention. In line with the prevalence of CNN network architectures, mainstream crowd counting schemes have transitioned from detection {\cite{ref1,ref2,ref3,ref4}} and regression {\cite{ref5,ref6,ref7}} to density estimation strategy {\cite{ref8,ref9,ref10}}, where each pixel represents the number of head counts at the corresponding location, thereby reducing the counting task to an accumulation of probabilities. Ideally this is possible, but in real dense scenes with varying head scales and uneven density distribution, the density map cannot clearly reflect the information. In addition, heavily obscured areas are extremely similar to the complex background, further exacerbating the error. Hence, a robust counting model requires strong generalization ability to external disturbances such as noisy background, scale variation, mutual occlusion, perspective distortion, etc.

\begin{figure}[!t]
\centering
\includegraphics[scale=0.3]{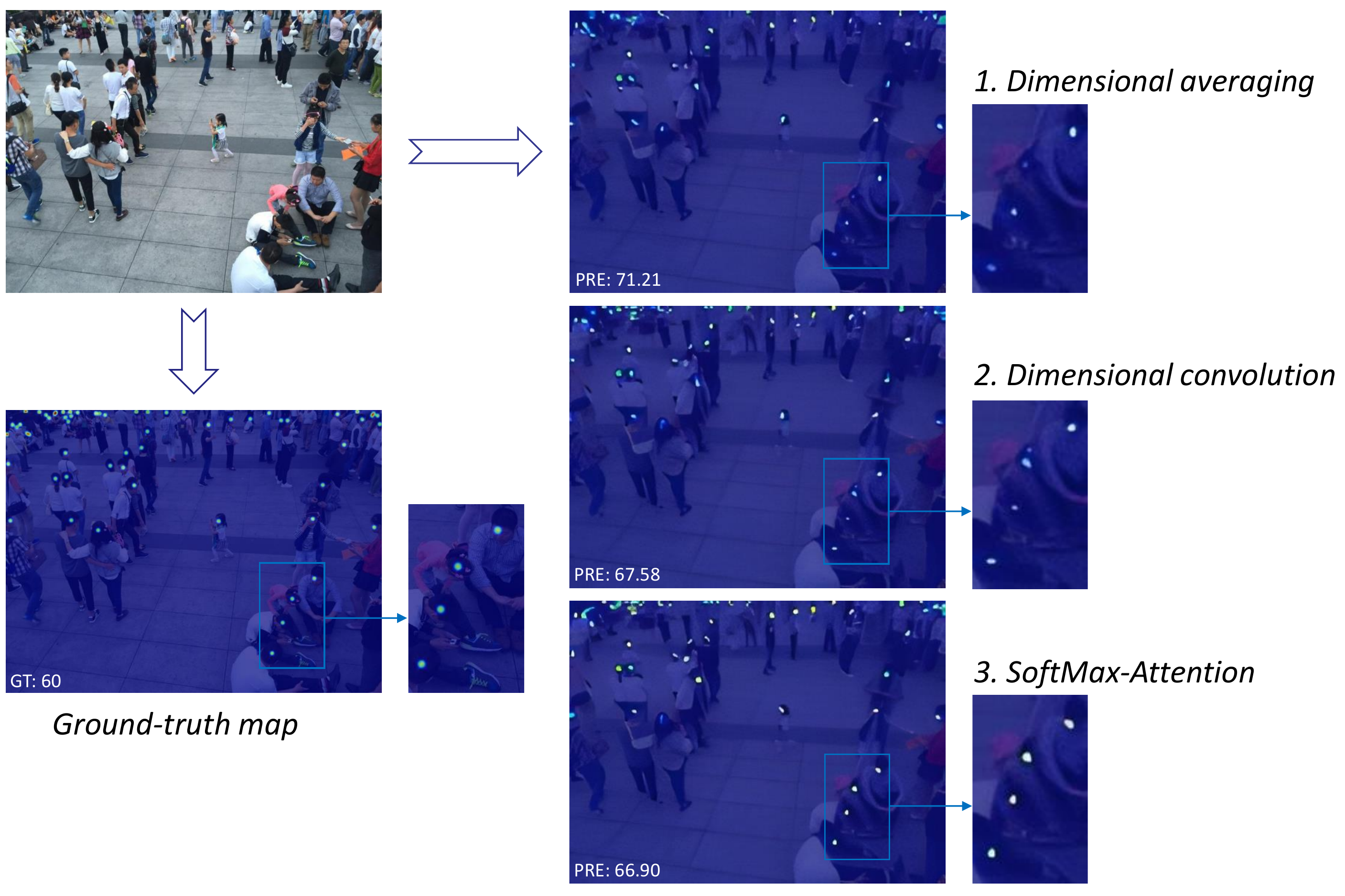}
\caption{  Comparison of the final predicted  density maps under different dimensionality reduction methods for high-dimensional attention map. } 
\label{Figure_1}
\end{figure}

The attention mechanism emphasizes biasing the computational focus  towards areas where the signal response is more pronounced, rather than processing the entire image indiscriminately {\cite{ref11}}. It has been widely used in  neural networks for a long time with impressive records {\cite{ref12}}, yet  there is still room for optimization. In a nutshell, we note two major flaws in the flow of loss calculation. First, the current attention label requires the involvement of artificial threshold {\cite{ ref12}}, which is less robust. Specifically, the point annotations are first processed by Gaussian function to get the density map, and the value of each position represents the count of heads here. If this value is greater than the threshold, the corresponding position in the attention label is considered as the head region, otherwise it is the background {\cite{ ref12}}. It can be seen that the quality of labels depends largely on the selection of thresholds, which are not universal in different crowd scenes and cannot guarantee the correctness of attention itself, let alone provide authoritative guidance for feature learning. Additionally, we can only make 1D labels, but the attention map in the network waiting for  loss calculation is necessarily high-dimensional. If convolutional dimensionality reduction is adopted, the source of loss becomes both the  attention itself and the convolutional parameters introduced by the dimensionality reduction. Thus the efficiency of attentional supervision is discounted and network convergence slowly. And if dimensional averaging is utilized, it means that each channel has the same weight, leading to the neutralization of high and low response features, which runs counter to the essence of the attention mechanism.

Based on the above analysis, in this paper, we devise the Indirect-Instant Attention Optimization (IIAO) module, as shown in Figure \textcolor{red}{\ref{Figure_3}}. It has two main components: 1) The submodule Adaptive Scale Pyramid (ASP), which follows the feature pyramid paradigm {\cite{ ref13}} to alleviate the scale variability troubles; 2) The submodule SoftMax Attention (SMA), which transforms high-dimensional attention map into a  one-dimensional feature map in the mathematical sense, thus circumventing the attention optimization  puzzles mentioned above, while providing adaptive multi-scale feature fusion for ASP. As shown in Figure \textcolor{red}{\ref{Figure_1}}, this scheme  can show superior results in practice applications compared to dimensional averaging and convolutional dimensionality reduction. Soft Block is the core device of SMA, it has two residual inputs, which are high-dimensional attention map and high-dimensional feature map. The purpose of normalizing the former is to enable each pixel to learn the weights of each channel at that location, and the next step of element-wise multiplication with the high-dimensional feature map is the beginning of the transformation . By adding up the result just now in the direction of the channel, the mathematical meaning is re-flipped (compared to the sigmoid function): the pixel value represents the number of heads here again, instead of the probability of being in the center of a head. At this point, the high-dimensional attention is transformed into a one-dimensional feature map, whose labels are relatively easy to obtain and reliable, so this transformation is meaningful and feasible. This is an indirect way of correcting parameters in high-dimensional attention that are difficult to optimize. At the same time we recognize  that, due to the specificity of the transformation, the obtained feature data may be coarse and unsuitable for the final prediction map, so we place this processing in the network midcourse, occurring instantly at the moment just after the attention map has finished its supervisory role. Thus, it is an indirect-instant attention optimization.

Furthermore,we propose a tailored Region Correlation Loss (RCLoss) to handle  coarse data and penalize  continuous error-prone regions. Considering different density distributions in a certain  scene, we resort to sliding windows. Also, to ensure the continuity of information between sub-windows and to atomize error-prone regions as much as possible, we allow windows to overlap, where overlapping regions will be analyzed multiple times, but only extremely error-prone positions will be repeatedly penalized. RCLoss assists in the IIAO module, supporting its faster and more accurate convergence.

The contributions of this paper are highlighted below:
\begin{itemize}
\item{We propose the Indirect-instant attention Optimization (IIAO) module, which transforms the body of the loss calculation from a high-dimensional attention map to an ordinary 1D feature map, providing timely and reliable supervision for regression,  while contributing adaptive scale fusion service for multi-column architectures.}
\item{We propose a tailored Region Correlation Loss (RCLoss) to cooperate with the IIAO module, which reduces the local estimation error and accelerates model convergence.}
\item{The proposed scheme has shown excellent performance in several benchmark datasets, beating even many SOTA counting methods.}
\end{itemize}

\section{RELATED WORK}
\subsection{Multi-scale Feature Extraction Strategy}

This strategy emphasizes that targets at different scales need to be perceived by perceptual fields of different sizes, and is generally implemented using a multi-column convolutional architecture.

 MCNN {\cite{ref14}} first uses a three-column network architecture to extract multi-scale features to accommodate scale variations due to different camera angles. Inspired by {\cite{ref14}}, Switching-CNN {\cite{ref15}} retains the multi-column structure, but adds a classifier to select the best branch suitable for the current scale. In addition, for multi-scale feature fusion, it is locally adaptive rather than the global fixed strategy in {\cite{ref14}}. CSRNet {\cite{ref16}} adapts a dilated convolutional layer to increase the receptive field as an alternative to the pooling operations, but it tends to lead to  grid effects, further leading to local information loss. DADNet {\cite{ref17}} is dedicated to innovation in multi-scale feature fusion, which advocates the use of multi-column dilation convolution to effectively learn multi-scale visual contextual cues, containing both the multi-column idea in {\cite{ref14}} and the essence of dilation convolution in {\cite{ref16}}.   Unlike the above methods, AMSNet {\cite{ref18}} utilizes neural architecture search and introduces an end-to-end search encoder-decoder architecture to automatically design the network models. 

\subsection{Attention Mechanism Guidance Strategy}
Attention mechanism is activated by the sigmoid function, which directs the model to focus on regions where the signal response is obvious and suppresses background noise, thus acting as a top-level supervision.

SFANet \cite{ref12} trains attention as a separate task pathway, similar to W-Net {\cite{ref19}}. However, its attention label are generated by density map paired with artificial threshold, which is not robust enough to generalize to scenes with different densities. ASNet \cite{ref20} considers the density of different regions in an image varies widely, which leads to different counting performance, and thus proposes a density attention network. This method provides attention masks with different density levels for the convolutional extraction unit, which is an aid and also a supervision. RANet {\cite{ref21}} emphasizes the optimization of attention, using two modules to handle global attention and local attention separately, and then finally fusing them according to the interdependencies between features. PCC-Net {\cite{ref22}} encodes the global, local and pixel-level features of the crowd scene separately and processes them in separate tasks. One of the FBS modules is responsible for segmenting the head region and background to further eliminate erroneous estimates. This fuels the accuracy of the prediction,but the training cost is high. DADNet  \cite{ref17} innovatively proposes scale-aware dilated attention, which uses multi-column dilated convolution to obtain attention maps responsible for different receptive fields, and thus focuses on heads with different scales, reproducing scale-awareness in a new way.Also, considering the adaptability to complex issues of scale variations, feature fusion is handled by direct summation instead of concatenation.  Recognizing that it is often difficult to generate accurate attention maps directly, CFANet \cite{ref23} turns to a coarse-to-fine progressive attention mechanism through two branches, the crowd region recognizer (CRR) and the density level estimator (DLE). This mode both suppresses the influence of background to reduce misidentification and adaptively assigns attention weights to different density regions. SGANet \cite{ref24}, the underlying structure is similar to SFANet \cite{ref12}: the attention map is homologous to the initial feature map, and they both belong to the  dual path task. However, the former improves the quality of attention due to the introduction of the tuned inception-v3 \cite{ref25} as Encoder, which improves the model effect tremendously. Attention determines the authority of top-level supervision, and this method improves the decision-making ability of attention from the root.

\begin{figure*}[!t]
\centering
\includegraphics[scale=0.53]{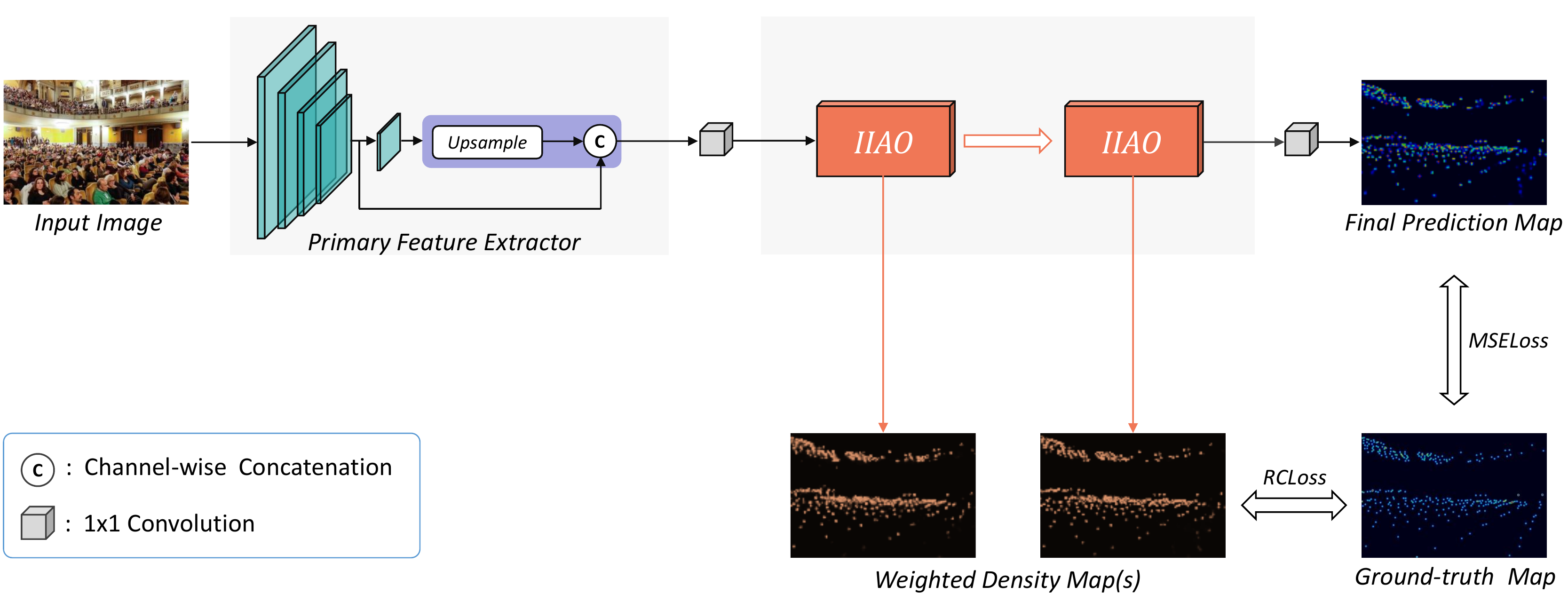}
\caption{Overall architecture of the proposed network. It first borrows the residual-connected VGG-16  as the backbone, and then stacks two IIAO modules consecutively with a convolution kernel at the front and back. The network generates two weighted density maps and a final prediction map, which calculate the loss using RCLoss and MSELoss, respectively.}
\label{Figure_2}
\end{figure*}

\subsection{  Practical Application of Softmax Function }

SASNet {\cite{ref26}} applies softmax to the traditional convolutional neural network, using it to rescale the attention weights, multiply them with the high-dimensional feature maps, and select the highest scoring hierarchical feature map among the resulting feature fusion maps. Transformer {\cite{ref27}} uses softmax to guarantee the non-negativity of the matrix and local attention amplification. However, its time complexity is the square of the sequence length, resulting in excessive overhead. cosformer{\cite{ref28}} replaces non-decomposable nonlinear softmax operations with linear operations with a decomposable nonlinear reweighting mechanism, which not only achieves comparable or better performance than softmax-attention across a range of tasks, but also has linear space and time complexity. {\cite{ref29}} introduces softmax into network pruning. Softmax-attention channel pruning consists of training, pruning, and fine-tuning steps. In the training step, it trains the network to be pruned; in the pruning step, it uses softmax to determine the importance of each channel and remove the relatively unimportant channels; in the fine-tuning step, the pruned model is trained with the same epochs used in the training step above {\cite{ref29}}.

\section{METHOD}

This paper aims to establish a crowd counting framework that is suitable for dense scenes. The architecture of the proposed method is illustrated in Figure \textcolor{red}{\ref{Figure_2}}. It includes a primary feature extractor taken from the VGG-16 {\cite{ref30}} model as the backbone, an additional convolutional layer for adjusting the dimensionality, two consecutive IIAO modules, and another convolutional layer for the final prediction map regression. In this chapter, we first briefly describe the backbone, and then focus on the proposed IIAO module and the RCLoss loss function.

\subsection{Primary Feature Extractor} \label{Section_3.1}

Similar to previous work {\cite{ref12}}, we place the front thirteen convolutional layers of VGG-16 {\cite{ref30}} with four pooling layers in an encoder  for extracting low-level  features, such as edges and textures. Its  output  map is then upsampled spatially by a factor of $2$ using bilinear interpolation. Immediately afterwards, the upsampled map is merged with the feature map obtained by the third convolution via channel-wise concatenation. Finally, the merged feature map is convolved through a $1 \times 1$  layer  to obtain $ F_{in} $ as the input of the first IIAO module, where  convolution is used to reduce the aliasing effect due to  upsampling {\cite{ref31}}. Hence, the generated $ F_{in} $ spatial size is $8$ times smaller than the original input. Note that the resolution of $ F_{in} $ no longer changes during all subsequent processing before the last single convolutional layer.

\subsection{ Indirect-Instant Attention Optimization Module}

\begin{figure*}[!t]
\centering
\includegraphics[scale=0.53]{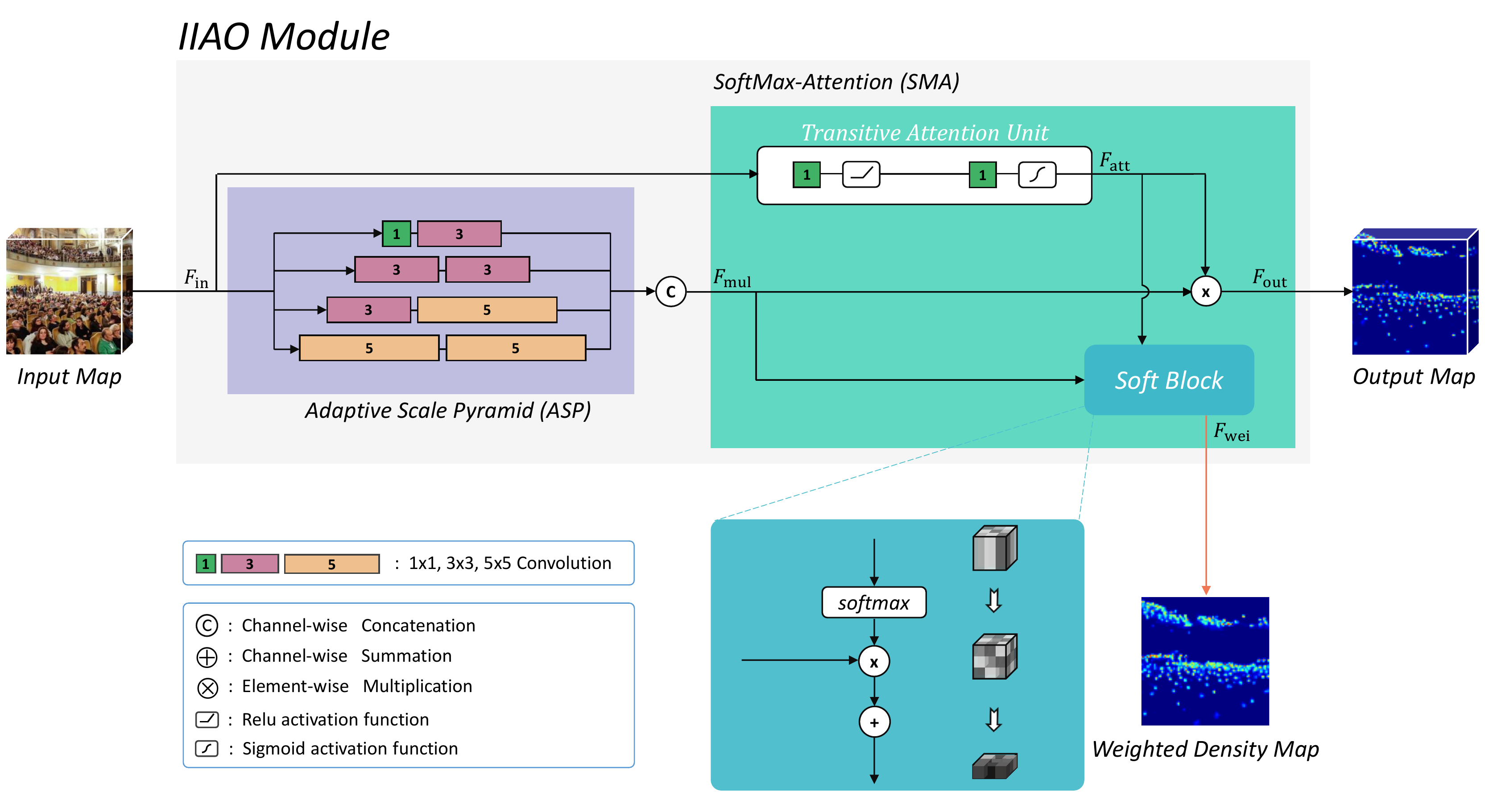}
\caption{Structural details of the proposed IIAO module, which contains an Adaptive Scale Pyramid (ASP) submodule and a SoftMax-Attention (SMA) submodule, where the generated one-dimensional weighted density map jumps out instantly for loss calculation, while the high-dimensional output map keeps the dimensionality continues to pass backwards.
}
\label{Figure_3}
\end{figure*}

As shown in Figure \textcolor{red}{\ref{Figure_3}}, the Indirect-Instant Attention Optimization (IIAO) module consists of two main components: the Adaptive Scale Pyramid (ASP) submodule and the SoftMax-Attention (SMA) submodule. As mentioned in Section \textcolor{red}{\ref{Section_3.1}}, $ F_{in} \in \mathbb{R}^{C \times H \times W} $ is an input  to IIAO module, where $C$ denotes the number of channels, and $H$ and $W$ represent the height and width, respectively, which are 8 times smaller than those of the original image. $F_{in}$ will produce two different types of feature maps each time it passes through the IIAO module: $F_{out}$ continues to pass backward; while $F_{wei}$  fuses attention directly with the ground-truth map for loss calculation. 

{\bf Adaptive Scale Pyramid (ASP) submodule} utilizes a multi-column architecture to acquire multi-scale features, while automatically completing the multi-scale feature fusion task at the channel-level along with the generation of $F_{wei}$ (see the next submodule for details), alleviating the limitation where the receptive field of each branch is fixed within a certain range. In detail, to reduce the parameter overhead, we conduct a $1\times1$ convolution filter at the beginning of ASP to compress the dimensionality of $F_{in}$ to $1/4$, and then expand it into four branches, each containing two of the three sizes of convolution filters $1\times1$, $3\times3$, and $5\times5$. In accordance with the design pattern of feature pyramid {\cite{ ref13}}, the further down the branch, the larger the perceptual field, so as to build a multi-scale perceptron to obtain contextual information. In each branch, the first convolution filter reduces the dimension by another factor of $4$ , and after collating the feature information, the second convolution filter recovers it, at which point the dimension of $F_{in}$ in each branch is $[C/4, H, W]$. Along these lines, the $F_{mul}$ generated  by the ASP submodule has exactly the same dimensions as the original $F_{in}$. Since there is no message passing or parameter sharing in  ASP , $F_{mul}$ at this time just incorporates features from multiple scales without fusion or selection, which is essentially an immature aggregate. We defer this work to the SMA submodule, whose process of generating $F_{wei}$  also involves  the model learning the dynamic weights of each location  in $F_{mul}$ online for all its channels, and finally using this weight, which represents the critical degree of information, to complete the adaptive multi-scale feature fusion task.

{\bf SoftMax-Attention (SMA) submodule} receives $F_{in}$ from the residual connection, and the Transitive Attention Unit provides it  contextual attention to obtain $F_{att}$. Specifically,  to facilitate arithmetic and extract  feature information of diverse levels, the Transitive Attention Unit first uses  $1\times1$ convolution filter with ReLU activation function  to reduce the channels of $F_{in}$, and obtain $ F_{in} \in \mathbb{R}^{C/r \times H \times W} $, where $r$ is the hyperparameter, specifying the reduction ratio. Then use another $1\times1$ convolution filter  to restore it and modulate with the sigmoid function to get the global context attention, denoted by $F_{att} \in \mathbb{R}^{C \times H \times W}$. At this node, $F_{att}$ splits into two paths,  {\bf the first path} replicates the traditional attention mechanism {\cite{ref12}}, applying element-wise multiplication to $F_{att}$ and $F_{mul}$, thus playing a supervisory role to enhance the key information in $F_{mul}$ and suppress its background noise. The result of multiplying the two is represented by $F_{out}$, which is still identical to the shape of $F_{in}$, and is then transmitted to the subsequent network sections. 
\begin{equation}
F_{out} = F_{att} \otimes F_{mul}
\end{equation}

After the last IIAO module, $F_{out}$ is regressed by a $1\times1$ convolution filter to generate the final prediction map $F_{pre}$.

Crucially, if we pursue timely and reliable attentional supervision, we should start the loss and gradient computation immediately after multiplying $F_{att}$ by $F_{mul}$, while keeping all its parameters unchanged. However, at this time, $F_{att}$ is generally in a high-dimensional state, and the corresponding pseudo-label can only be one-dimensional. Therefore, before calculating the loss, we must first reduce the dimensionality of $F_{att}$. To average it would mean that all feature layers have the same weight,  the key information and environmental noise would be neutralized, which is contrary to the original intention of the attention mechanism, whereas if convolutional dimensionality reduction is used, additional learnable parameters will be introduced, which makes it impossible to know whether the effect of the model stems from the attention supervision or the convolution learning during the process of dimensionality reduction.

Based on the above analysis, we propose a SoftMax-Attention strategy to optimize the loss calculation process of attention in {\bf the second path}. This approach removes pseudo-labeling and does not introduce additional learnable parameters, enabling loss and gradient computation to correct the kernel parameters immediately after acting as a supervisor. We place this device halfway through the network multiple times to ensure that the convergence of the network is always on track.   To expand, after each $F_{att}$ is taken from Transitive Attention Unit,   it is transformed into a probability distribution between $[0,1]$ in the channel direction using the softmax function, with the aim of learning the dynamic weight of the feature expressed by each pixel in $F_{mul}$ across all channels at that location. Then, $F_{mul}$ is multiplied by $F_{att}$, noting that the result is different from $F_{out}$. All channels are summed vertically to obtain the weighted density map of fused attention to features, denoted by $F_{wei} \in \mathbb{R}^{1 \times H \times W}$. 

Up to this point, the body of the attention loss calculation is mathematically transformed from a high-dimensional attention map to an ordinary one-dimensional feature map, $F_{wei}$, because what each of its pixel values represents is, again, the number of heads at that location. We know that 1D feature map labels are easy to produce and relatively reliable, so this special transformation is feasible and meaningful. This is an indirect way of correcting parameters in high-dimensional attention that are difficult to optimize; at the same time, this correction occurs just after the attention map has completed its supervisory role, so it is also an instant way. Combined, this is called Indirect-Instant attention.

\begin{equation}
F_{wei_{i,j}} \! = \! \sum _{k=1}^{C}(SoftMax(F_{att}) \otimes F_{mul})_{k,i,j} \; 
\begin{cases}
\! 1 \!\! \le \!\! i \!\! \le \!\! H \\ 
\! 1 \!\! \le \!\! j \!\! \le \!\! W
\end{cases}
\end{equation}

Another thing to note is that this device allows the network to learn the weights of each channel under each branch, so the SoftMax-Attention acts on different feature branches. And different branches have different perceptual capabilities for features of different scales, so the process to get $F_{wei}$ also completes the task of multi-scale feature fusion for the ASP submodule. 

To summarize, the SoftMax-Attention strategy enables the attention map to pre-emptively complete loss and gradient computation in an indirect way midway through the network, thus addressing  two defects in the attention loss calculation, while it provides adaptive multi-scale feature fusion service for the ASP submodule. So this is a simple but efficient approach.  Ablation experiments demonstrate that the SoftMax-Attention strategy achieves better results than averaging or convolutional dimensionality reduction methods.

SASNet {\cite{ref26}} also borrows the softmax algorithm, which utilizes VGG-16 {\cite{ref30}} network regression to obtain 5 layers of Confidence Head and Density Head. The two are then upsampled to the same size and each is concatenated to obtain Confidence Maps and Density Maps. Finally, the former is softmaxed and multiplied with the latter, and then summed by dimensional direction to 
get the final prediction map. It can be seen that the basic action is similar, but we note that the features obtained by this treatment may not be smooth enough,  and saturated training on it  is not the only choice, so instead of using it as the final estimates , we place it in the middle  of the network and target the optimization by introducing RCLoss that focus on error-prone regions. In addition, the concatenation operation requires undifferentiated upsampling of multilayer feature maps with a maximum ratio of up to 16, which severely blurs the information at deeper levels and thus puts it at a disadvantage in the scale selection stage. In contrast to rough processing, we focus on the particular advantages of softmax dimensionality reduction. Stacking IIAO modules to progressively improve attentional reliability and guiding feature pyramid  regression to learn prediction maps for high ratings. Also, feeding high-resolution patches alleviates the problem of feature extinction in overly deep network.

\subsection{Loss Function} \label{Section_3.3}

{\bf Euclidean Distance.} The vast majority of researchers tend to choose the euclidean distance to measure the pixel error between the final prediction map and the ground-truth density map, which is defined as follows:

\begin{equation}
L_{pre}=\frac{1}{N}\sum_{i=1}^N{\left\| P\left( X_i;\varTheta \right) -G_{i}^{GT} \right\| _{2}^{2}}
\end{equation}
Where $N$ is the number of images in a training batch, $X_{i}$ denotes the current training image, $\varTheta$ is a set of learnable parameters in the network, so $ P\left( X_i;\varTheta \right)$ represents the prediction map for it, and $G_{i}^{GT}$ refers to its ground-truth density map.

\begin{figure*}[!t]
\centering
\includegraphics[scale=0.53]{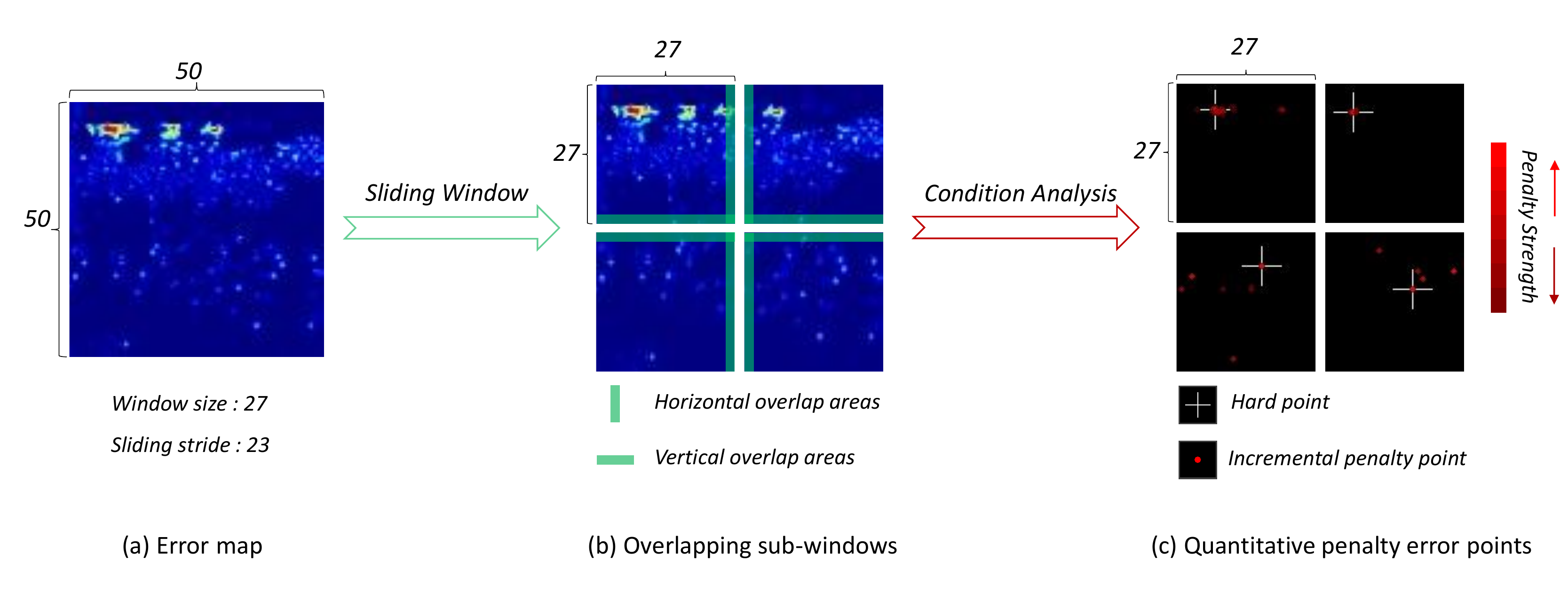}
\caption{RCLoss function computational flow. (a) Error map of the model predictions. (b) Sub-windows containing overlapping areas, designed to prevent heads from being cut to affect the error analysis. (c) Each sub-window performs error analysis separately. First, define the position with the largest error value as the hard point; then, determine whether the error at the remaining positions is higher than the product of the hard point value and $threshold$. If so, apply an incremental penalty, where the greater the error is, the greater the penalty increment it receives. Otherwise, simply take the square value of the original error as the penalty.}
\label{Figure_4}
\end{figure*}

{\bf Regional Correlation Loss.} Realistic scenes with different densities of people are unevenly distributed, and there may be multiple spatially uncorrelated error-prone regions in a single image, but MSELoss assumes that the pixels are isolated and independent from each other, and spatial correlation cannot be guaranteed. In addition,  although  the $F_{wei}$ generated by Soft Block is mathematically equivalent to the ordinary density map,  the internal data are coarser due to the specificity of the transformation  method. To cope with above problem, we innovatively propose the Regional Correlation Loss function (RCLoss) and apply it to the IIAO module, the pairing of the two can optimize attention more effective. At the end of the network, the more general MSELoss is used to act on the final prediction map, which is actually a review of the effectiveness of the IIAO module paired with RCLoss.  The computational flow of RCLoss is shown in Figure \textcolor{red}{\ref{Figure_4}}.

RCLoss focuses on using the $F_{wei}$ generated by the SMA submodule, subtracts it from the ground-truth map and obtains the absolute value, where the result is represented by the Error map. Set the sliding window to traverse the Error map, find the pixel point with a large error in each of the obtained sub-windows, and apply an incremental penalty based on the error value of the position itself, \textit{i.e.} , the strength of the penalty is determined by the degree of error in that position. To ensure information continuity between sub-windows, we set the sliding stride slightly smaller than the sliding window size. In this way, the resulting overlapping regions will be analyzed multiple times, but only extremely error-prone positions will be repeatedly penalized, and low-sensitivity areas are simply ignored. 

In detail, the Error map, denoted by $E \in \mathbb{R}^{1 \times H \times W}$, is obtained first.

\begin{equation}
E=\left| F_{wei}-G^{GT} \right|
\end{equation}
Assuming that the size and stride of the sliding window are $k$ and $s$, respectively, then the maximum sliding times of the window in the horizontal and vertical directions can be calculated, which are represented by $R_{max}$ and $D_{max}$, respectively.

\begin{equation}
R_{\max}\,\,=\,\,\lfloor \frac{W-k}{s} \rfloor +1, D_{\max}\,\,=\,\,\lfloor \frac{H-k}{s} \rfloor +1
\end{equation}

Each sub-window has its pixel point with the largest value of its error, called the hard point, whose value is $ MAX_{r,d}, \{1 \! \le \! r \! \le \! R_{max}, 1 \! \le \! d \! \le \! D_{max}\} $. By analyzing each sub-window in turn, penalizing positions where the error value is close to $MAX_{r,d}$, the amount of penalty is determined by the constant $threshold$, and the strength of the penalty is related to the degree of deviation from its own prediction. For those  points where the error value is within the tolerance range, MSELoss is used as their loss . The specific RCLoss algorithm is shown below:
\begin{equation*}
\begin{aligned}
EP: E_{i,j} > MAX_{r,d}* threshold \\
ET: E_{i,j} \leq MAX_{r,d}* threshold 
\end{aligned}
\end{equation*}

\begin{equation}\label{Formula_6}
\!\!Loss_{r,d}\!= \!
\begin{cases}
\sum\limits_{i=r*s+1}^{r*s+k}{\sum\limits_{j=d*s+1}^{d*s+k}{\!\!(\frac{E_{i,j}}{1+e^{-E_{i,j}}}\!+\!E_{i,j}) ^2}}&\! if \; EP\\
\sum\limits_{i=r*s+1}^{r*s+k}{\sum\limits_{j=d*s+1}^{d*s+k}{ \qquad {E_{i,j}}^2}}& \! if \; ET\\
\end{cases}
\end{equation}
$Loss_{r,d}$ represents the total loss of each sub-window, and $i$ and $j$ are iterators of its width and height, respectively. Within each sub-window, first determine the hard point and calculate its error value $MAX_{r,d}$; then, search for error-prone points whose error value is greater than the product of $threshold$ and $MAX_{r,d}$;  and finally calculate the loss together with the rest of the ordinary positions according to Equation \eqref{Formula_6} . $EP$ and $ET$ represent error-prone and error-tolerant points , respectively.

All reachable sub-window losses in a training batch are accumulated  to give the final RCLoss:

\begin{equation}
L_{wei}=\frac{1}{N}\sum_{i=1}^N{\sum_{r=1}^{R_{\max}}{\sum_{d=1}^{D_{\max}}{Loss_{r,d}}}}
\end{equation}

{\bf Final Loss Function.} The entire network produces three density maps of the same size, including two $F_{wei}$s output from the IIAO module and a final prediction map $F_{pre}$.  By weighting the different tasks, the unified objective function required by training can be formulated as:

\begin{equation}
L=\lambda L_{wei1}+\lambda L_{wei2}+\gamma L_{pre}
\end{equation}
where $\lambda$ and $\gamma$ are the weight terms of the two loss functions. Both can be set to fixed values for experiments with all datasets, showing an excellent generalization ability.

SASNet{\cite{ref26}} also  mentions the concept of hard pixel, but it is quite different from this paper, which can be summarized in three aspects. \textbf{1)}  It cuts the prediction map into four parts, selects the one  with the largest total error, and then recursively cuts and compares until the pixel with the largest error in the entire prediction map is determined, which is called hard pixel. By contrast, we consider that the probability of large-scale human heads appearing on the dividing line is high, and forcible cutting will cause irreversible damage to potential targets and affect the prediction of the network. We therefore propose sliding windows with overlapping regions and perform ablation experiments for the overlap length, which is finally determined to be $8$. Since the final prediction map has the same \textit{8X} perceptual field, the mapping back to the original map is a $64 \times 64$ region, which is close to the size of a larger human head. \textbf{2)} The number of penalty objects is different. Our hard point is not the only object to be punished, but plays a more important role as a reference. Each of the four sub-windows has its own hard point, which indicates the position with the largest prediction deviation in the current sub-window. Suppose its value is MAX, search for points in this sub-window whose error value is close to MAX, treat them as penalty objects as well, and whether they are close or not is evaluated by threshold. The purpose of this move  is to take into account the uneven distribution of head density, where errors may be concentrated in a certain area. Restricting to hard pixel may have limited effect. \textbf{3)} The intensity of penalty varies among different penalty objects; after all, their degree of prediction deviation is different. The hard point, as the biggest troublemaker, undoubtedly receives the most attention; for its affiliated penalty points, the penalty intensity is softened with the decrease of the error . See Equation \eqref{Formula_6} for  specific rules.

\section{Training}

This chapter introduces the specific training details in terms of data preprocessing, label generation, and hyperparameter setting.

\subsection{ Data Pre-processing}
In the training phase, we randomly crop a $400 \times 400$ patch and flip it horizontally with a probability of $0.5$.  For the ShanghaiTech Part\_A{\cite{ref14}}  and UCF-QNRF{\cite{ref32}} datasets containing grey images, we change the color images to grey with a probability of $0.1$.  During the test phase, for the datasets containing extremely large resolution, \textit{i.e.}, UCF-QNRF{\cite{ref32}}, JHU-Crowd++{\cite{ref33}} and NWPU-Crowd{\cite{ref34}},we scale images with side lengths greater than $5000$ to $4/5$ of their original size.

\subsection{ Label Generation }

As in previous work {\cite{ref12}}, we blur each head annotation with a Gaussian kernel to generate training labels. In detail, for crowd-sparse datasets, such as ShanghaiTech Part\_B {\cite{ref14}}, we use fixed-size kernels to generate ground-truth, while for other datasets with relatively dense scenes, geometric adaptive kernel based on the nearest neighbor algorithm are utilized.

\subsection{ Hyperparameter Setting }
Except for Primary Feature Extractor, the parameters of the subsequent layers are randomly initialized by a Gaussian distribution with a mean of $0$ and a standard deviation of $0.01$. The  $r$ in  Transitive Attention Unit is set to $16$. For training details, we choose the Adam {\cite{ref35}} optimizer to retrain the model, with an initial learning rate of 1E-4, halved every 100 rounds.  Weight items $\lambda$ and $\gamma$ are set to $1.5$ and $0.5$, respectively, in the training of all datasets. 

\section{Experiments}

We demonstrate the effectiveness of the proposed method on six official datasets: ShanghaiTech  {\cite{ref14}}, UCF\_CC\_50 {\cite{ref6}}, UCF-QNRF {\cite{ref32}}  JHU-Crowd++ {\cite{ref33}}  and NWPU-Crowd{\cite{ref34}}.

\subsection{ Evaluation Metrics}

There are two mainstream metrics for evaluating the performance in crowd counting task: Mean Absolute Error (MAE) and Mean Squared Error (MSE). They are defined as follows:

\begin{equation}
MAE=\frac{1}{N}\sum_{i=1}^N{\left| P_i-G_i \right|}
\end{equation}

\begin{equation}
MSE=\sqrt{\frac{1}{N}\sum_{i=1}^N{\left| P_i-G_i \right|^2}}
\end{equation}
where $N$ is the number of images in the test set, and $P_{i}$ and $g_{i}$ are the predicted number of targets in the $i$-th image and its corresponding ground-truth number, respectively.

\subsection{ Comparisons and Analysis }

{\bf ShanghaiTech } dataset is composed of two parts. Part\_A  contains 482 images, sourced from the web, with rich scenes and variable scales, and the number of images used for training and test are 300 and 182, respectively; Part\_B  contains 716 images, collected from Shanghai streets, with  relatively sparse  distribution and a fixed image size of $1024 \times 768$, further divided into a training set containing 400 images and a test set containing 316 images. The experimental results  are shown in Table \textcolor{red}{\ref{Table_1}}, which indicates that our method  achieves  the desired level in both dense and sparse scenes.

\begin{table*}
\caption{\centering Comparison with state-of-the-art methods on four challenging datasets. The best performance is indicated by {\bf bold} and the second best is {\ul underlined}. \label{Table_1}}  
\centering
\begin{tabular}{|l|l|cc|cc|cc|cc|}
\hline
\multirow{2}{*}{Methods} & \multicolumn{1}{c|}{\multirow{2}{*}{Venue}} & \multicolumn{2}{c|}{Part\_A}    & \multicolumn{2}{c|}{Part\_B} & \multicolumn{2}{c|}{UCF\_CC\_50}  & \multicolumn{2}{c|}{UCF-QNRF}    \\ \cline{3-10} 
                         & \multicolumn{1}{c|}{}                       & MAE            & MSE            & MAE           & MSE          & MAE             & MSE             & MAE            & MSE             \\ \hline
MCNN {\cite{ref14}}      & CVPR 16                                     & 110.2          & 173.2          & 26.4          & 41.3         & 377.6           & 509.1           & 277.0          & 426.0           \\ \hline
Deem {\cite{ref36}}      & T-CSVT 19                                   & -              & -              & 8.09          & 12.98        & 253.4           & 364.4           & -              & -               \\
PCC-Net {\cite{ref22}}   & T-CSVT 19                                   & 73.5           & 124.0          & 11.0          & 19.0         & 183.2           & 260.1           & 148.7          & 247.3           \\
DADNet {\cite{ref17}}                  & ACM-MM 19                                   & 64.2           & 99.9           & 8.8           & 13.5         & 285.5           & 389.7           & 113.2          & 189.4           \\
BL {\cite{ref37}}        & ICCV 19                                     & 62.8           & 101.8          & 7.7           & 12.7         & 229.3           & 308.2           & 88.7           & 154.8           \\
RANet {\cite{ref21}}     & ICCV 19                                     & 59.4           & 102.0          & 7.9           & 12.9         & 239.8           & 319.4           & 111.0          & 190.0           \\ \hline
ZoomCount {\cite{ref38}} & T-CSVT 20                                   & 66.0           & 97.5           & -             & -            & -               & -               & 128.0          & 201.0           \\
DM-Count {\cite{ref39}}  & NeurIPS 20                                  & 59.7           & 95.7           & 7.4           & 11.8         & 211.0           & 291.5           & 85.6           & 148.3           \\
ASNet {\cite{ref20}}                   & CVPR 20                                     & 57.78          & 90.13          & -             & -            & 174.84          & 251.63          & 91.59          & 159.71          \\
LibraNet {\cite{ref40}}  & ECCV 20                                     & 55.9           & 97.1           & 7.3           & 11.3         & 181.2           & 262.2           & 88.1           & 143.7           \\
AMSNet {\cite{ref18}}    & ECCV 20                                     & 56.7           & 93.4           & 6.7           & {\ul 10.2}   & 208.4           & 297.3           & 101.8          & 163.2           \\ \hline
DCANet {\cite{ref41}}    & T-CSVT 21                                   & 59.2           & 94.4           & 7.8           & 12.3         & 183.2           & 260.1           & 90.1           & 150.4           \\
CFANet {\cite{ref23}}                  & WACV 21                                     & 56.1           & {\ul 89.6}     & {\ul 6.5}     & {\ul 10.2}   & 203.6           & 287.3           & 89.0           & 152.3           \\
DKPNet {\cite{ref42}}    & ICCV 21                                     & 55.6           & 91.0           & 6.6           & 10.9         & -               & -               & 81.4           & 147.2           \\
GLoss {\cite{ref43}}     & CVPR 21                                     & 61.3           & 95.4           & 7.3           & 11.7         & -               & -               & 84.3           & 147.5           \\
D2C {\cite{ref44}}                     & TIP 21                                      & 59.6           & 100.7          & 6.7           & 10.7         & 221.5           & 300.7           & 84.8           & 145.6           \\
SASNet {\cite{ref26}}    & AAAI 21                                     & \textbf{53.59}    & \textbf{88.38} & \textbf{6.35} & \textbf{9.9} & {\ul 161.4}     & {\ul 234.46}    & 85.2           & 147.3           \\
TopoCount {\cite{ref45}} & AAAI 21                                     & 61.2           & 104.6          & 7.8           & 13.7         & 184.1           & 258.3           & 89             & 159             \\ \hline
SGANet {\cite{ref24}}                  & TITS 22                                     & 57.6           & 101.1          & 6.6           & {\ul 10.2}   & 221.9           & 289.8           & 87.6           & 152.5           \\
MAN {\cite{ref46}}       & CVPR 22                                     & 56.8           & 90.3           & -             & -            & -               & -               & {\ul 77.3}     & {\ul 131.5}     \\ \hline \hline
Ours                     & -                                           & {\ul 54.37} & 92.76          & 6.94          & 10.80        & \textbf{147.96} & \textbf{202.12} & \textbf{74.01} & \textbf{128.87} \\ \hline 
\end{tabular}
\end{table*}

\begin{table*}
\caption{\centering  Comparison with state-of-the-art methods on the JHU-Crowd++ (val set) dataset. "Low", "Medium" and "High"  respectively indicates three categories based on different ranges: [0,50], (50,500) and \textgreater 500. The best performance is indicated by {\bf bold} and the second best is {\ul underlined}. \label{Table_2}}
\centering
\begin{tabular}{|l|l|cccccccc|}
\hline
\multirow{3}{*}{Method}    & \multirow{3}{*}{Venue} & \multicolumn{8}{c|}{Val set}                                                                                                                                                                          \\ \cline{3-10} 
                           &                        & \multicolumn{2}{c|}{Low}                           & \multicolumn{2}{c|}{Medium}                          & \multicolumn{2}{c|}{High}                              & \multicolumn{2}{c|}{Overall}     \\ \cline{3-10} 
                           &                        & MAE           & \multicolumn{1}{c|}{MSE}          & MAE            & \multicolumn{1}{c|}{MSE}           & MAE             & \multicolumn{1}{c|}{MSE}            & MAE            & MSE            \\ \hline
MCNN {\cite{ref14}}              & CVPR 16                & 90.6          & \multicolumn{1}{c|}{202.9}         & 125.3          & \multicolumn{1}{c|}{259.5}          & 494.9           & \multicolumn{1}{c|}{856.0}             & 160.6          & 377.7           \\ \hline
CSRNet {\cite{ref16}}            & CVPR 18                & 22.2          & \multicolumn{1}{c|}{40.0}            & 49.0             & \multicolumn{1}{c|}{99.5}           & 302.5           & \multicolumn{1}{c|}{669.5}           & 72.2           & 249.9           \\
SANet {\cite{ref47}}             & ECCV 18                & 13.6          & \multicolumn{1}{c|}{26.8}          & 50.4           & \multicolumn{1}{c|}{78.0}             & 397.8           & \multicolumn{1}{c|}{749.2}           & 82.1           & 272.6           \\ \hline
SFCN {\cite{ref48}}              & CVPR 19                & 11.8          & \multicolumn{1}{c|}{19.8}          & 39.3           & \multicolumn{1}{c|}{73.4}           & 297.3           & \multicolumn{1}{c|}{679.4}           & 62.9           & 247.5           \\
CAN {\cite{ref49}}               & CVPR 19                & 34.2          & \multicolumn{1}{c|}{69.5}          & 65.6           & \multicolumn{1}{c|}{115.3}          & 336.4           & \multicolumn{1}{c|}{{\ul 619.7}}     & 89.5           & 239.3           \\
DSSINet {\cite{ref50}}          & ICCV 19                & 50.3          & \multicolumn{1}{c|}{85.9}          & 82.4           & \multicolumn{1}{c|}{164.5}          & 436.6           & \multicolumn{1}{c|}{814.0}             & 116.6          & 317.4           \\
MBTTBF {\cite{ref51}}            & ICCV 19                & 23.3          & \multicolumn{1}{c|}{48.5}          & 53.2           & \multicolumn{1}{c|}{119.9}          & 294.5           & \multicolumn{1}{c|}{674.5}           & 73.8           & 256.8           \\
BL {\cite{ref37}}                & ICCV 19                & {\ul 6.9}           & \multicolumn{1}{c|}{{\ul 10.3}}          & 39.7           & \multicolumn{1}{c|}{85.2}           & 279.8           & \multicolumn{1}{c|}{620.4}           & 59.3           & {\ul 229.2}     \\ \hline
LSC-CNN {\cite{ref52}}           & PAMI 20                & \textbf{6.8}     & \multicolumn{1}{c|}{\textbf{10.3}}    & 39.2           & \multicolumn{1}{c|}{64.1}           & 504.7           & \multicolumn{1}{c|}{860.0}             & 87.3           & 309.0             \\
CG-DRCN-CC-VGG16 {\cite{ref33}}  & PAMI 20                & 17.1          & \multicolumn{1}{c|}{44.7}          & 40.8           & \multicolumn{1}{c|}{71.2}           & 317.4           & \multicolumn{1}{c|}{719.8}           & 67.9           & 262.1           \\
CG-DRCN-CC-Res101 {\cite{ref33}} & PAMI 20                & 11.7          & \multicolumn{1}{c|}{24.8}          & 35.2           & \multicolumn{1}{c|}{57.5}           & \textbf{273.9}     & \multicolumn{1}{c|}{676.8}           & {\ul 57.6}     & 244.4           \\ \hline
AutoScale {\cite{ref53}}         & IJCV 21                & 10.0            & \multicolumn{1}{c|}{15.3}          & {\ul 33.5}     & \multicolumn{1}{c|}{{\ul 54.2}}     & 351.7           & \multicolumn{1}{c|}{720.3}           & 65.7           & 258.9           \\ \hline  \hline

Ours                       & -                      & 8.24 & \multicolumn{1}{c|}{12.73} & \textbf{30.44} & \multicolumn{1}{c|}{\textbf{50.47}} & {\ul 276.57} & \multicolumn{1}{c|}{\textbf{589.73}} & \textbf{54.08} & \textbf{212.73} \\ \hline  
\end{tabular}
\end{table*}

{\bf UCF\_CC\_50 } dataset is a small but challenging dataset, containing 50 images collected from the internet. The number of people annotated in each image varies widely from 94 to 4,543. Due to the limited  samples, there is no official division between the training and test set, but rather a 5-fold cross-validation is suggested as a unified  test approach. We strictly follow this rule for our experiments, and the results are also  shown in Table \textcolor{red}{\ref{Table_1}}. It can be seen that our method is 8.33\% ahead of the SOTA level in terms of  MAE metric, and it is 13.8\% ahead in terms of MSE.  Thus, our method still performs well in extremely dense scenes.

{\bf UCF-QNRF } dataset  has  1535 images containing 1,251,642 annotation points. The training and test sets are composed of 1201 and 334 images, respectively. The density of annotated targets ranges from 49 to 12,865, and the average image size is $2013 \times 2902$, which poses an enormous challenge for  counting task. As shown in Table \textcolor{red}{\ref{Table_1}}, we achieve  improvements of 4.26\% and 2\% in terms of the MAE and MSE , respectively,  compared to the SOTA methods.

{\bf JHU-Crowd++ } dataset  has a more complex context than the above  datasets, with 4372 images and 1.51 million annotations. All images are divided into 2722 training images, 500 val images and 1600 test images. We follow the official description, classify val and test by density level, and compare them one by one. The results of both are shown in Tables \textcolor{red}{\ref{Table_2}} and \textcolor{red}{\ref{Table_3}}.  In the val set, both metrics in our \textit{}{Overall} and \textit{Medium} parts outperform the previous SOTA method by 6.11\%, 7.19\%, 9.13\%, and 6.88\%, respectively. In the \textit{High} part, the MAE trails that of the optimal method by 0.97\%, and  MSE leads by 4.84\%. More notably, for the test set, we  have made progress across the board in both \textit{High} and \textit{Medium} parts, leading the second place by 9.63\%, 5.82\%, 6.13\%, and 1.57\% in the two metrics, respectively. Also, for the \textit{Low} part, MSE surpasses the previous SOTA method by 27.58\%, which is a great improvement. Unfortunately, however, in the \textit{Overall} part, both metrics lag behind the MAN {\cite{ref46}}.

\begin{figure*}[!t]
\centering
\includegraphics[scale=0.53]{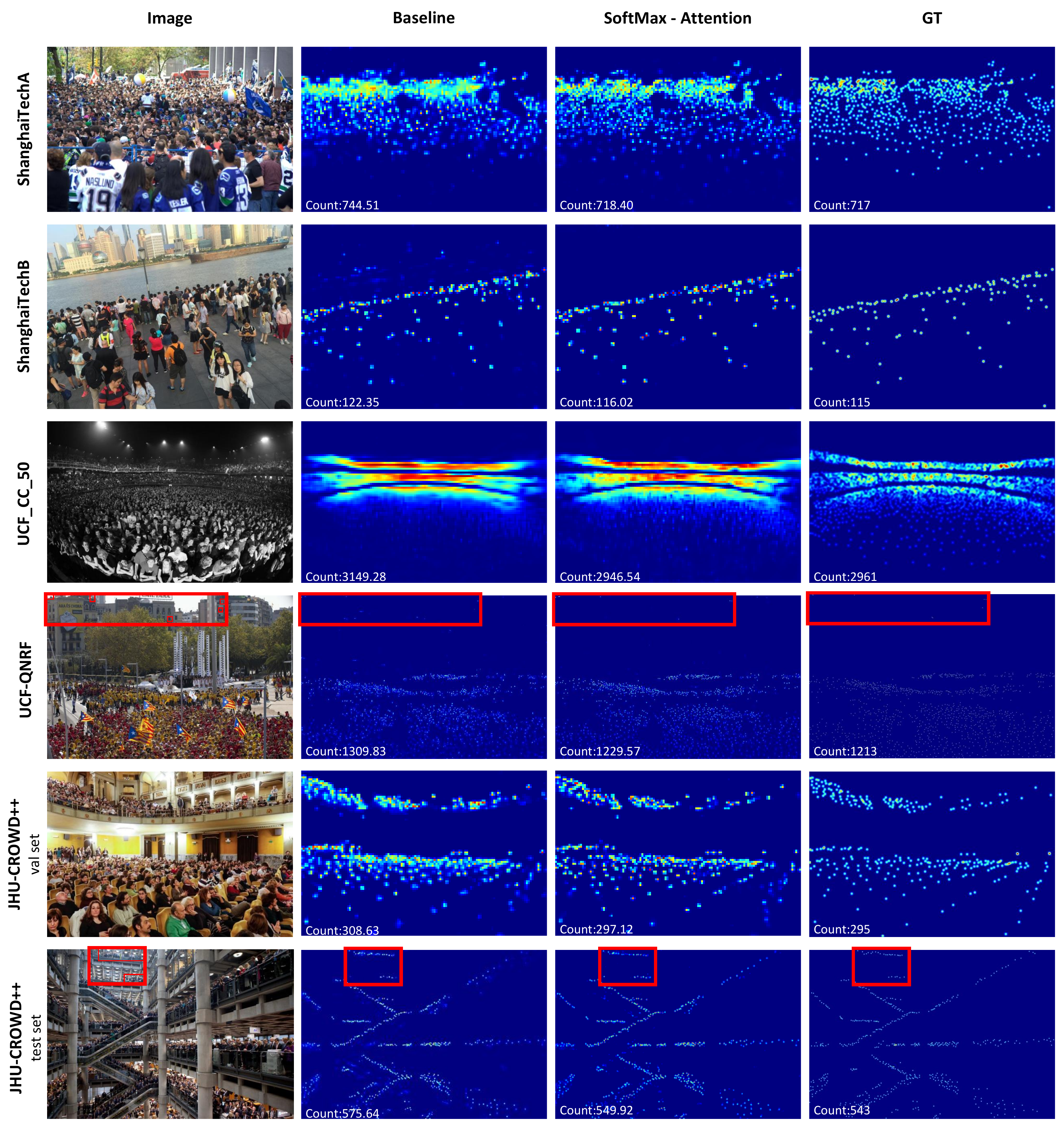}
\caption{A quantitative visualization showing the superiority of the SoftMax-Attention strategy in multiple datasets, where each row represents a dataset, from left to right: Input image, Baseline result, SoftMax-Attention result, Ground-truth map. \textit{Zoom in the figure for better viewing.}}
\label{Figure_5}
\end{figure*}

\begin{table*}
\caption{\centering Comparison with state-of-the-art methods on the JHU-Crowd++ (test set) dataset. "Low", "Medium" and "High"  respectively indicates three categories based on different ranges: [0,50], (50,500) and \textgreater 500. The best performance is indicated by {\bf bold} and the second best is {\ul underlined}. \label{Table_3}}
\centering
\begin{tabular}{|l|l|cccccccc|}
\hline
\multirow{3}{*}{Method} & \multirow{3}{*}{Venue} & \multicolumn{8}{c|}{Test set}                                                                                                                                                                        \\ \cline{3-10} 
                        &                        & \multicolumn{2}{c|}{Low}                            & \multicolumn{2}{c|}{Medium}                          & \multicolumn{2}{c|}{High}                              & \multicolumn{2}{c|}{Overall}   \\ \cline{3-10} 
                        &                        & MAE           & \multicolumn{1}{c|}{MSE}            & MAE            & \multicolumn{1}{c|}{MSE}            & MAE             & \multicolumn{1}{c|}{MSE}             & MAE           & MSE            \\ \hline
MCNN {\cite{ref14}}                    & CVPR 16                & 97.1          & \multicolumn{1}{c|}{192.3}          & 121.4          & \multicolumn{1}{c|}{191.3}          & 618.6           & \multicolumn{1}{c|}{1166.7}          & 188.9         & 483.4          \\ \hline
CSRNet {\cite{ref16}}                  & CVPR 18                & 27.1          & \multicolumn{1}{c|}{64.9}           & 43.9           & \multicolumn{1}{c|}{71.2}           & 356.2           & \multicolumn{1}{c|}{784.4}           & 85.9          & 309.2          \\
SANet {\cite{ref47}}                   & ECCV 18                & 17.3          & \multicolumn{1}{c|}{37.9}           & 46.8           & \multicolumn{1}{c|}{69.1}           & 397.9           & \multicolumn{1}{c|}{817.7}           & 91.1          & 320.4          \\ \hline
SFCN {\cite{ref48}}                    & CVPR 19                & 16.5          & \multicolumn{1}{c|}{55.7}           & 38.1           & \multicolumn{1}{c|}{59.8}           & 341.8           & \multicolumn{1}{c|}{758.8}           & 77.5          & 297.6          \\
CAN {\cite{ref49}}                     & CVPR 19                & 37.6          & \multicolumn{1}{c|}{78.8}           & 56.4           & \multicolumn{1}{c|}{86.2}           & 384.2           & \multicolumn{1}{c|}{789.0}           & 100.1         & 314.0          \\
DSSINet {\cite{ref50}}                & ICCV 19                & 53.6          & \multicolumn{1}{c|}{112.8}          & 70.3           & \multicolumn{1}{c|}{108.6}          & 525.5           & \multicolumn{1}{c|}{1047.4}          & 133.5         & 416.5          \\
MBTTBF {\cite{ref51}}                  & ICCV 19                & 19.2          & \multicolumn{1}{c|}{58.8}           & 41.6           & \multicolumn{1}{c|}{66.0}           & 352.2           & \multicolumn{1}{c|}{760.4}           & 81.8          & 299.1          \\
BL {\cite{ref37}}                      & ICCV 19                & \textbf{10.1}    & \multicolumn{1}{c|}{32.7}           & 34.2           & \multicolumn{1}{c|}{54.5}           & 352.0           & \multicolumn{1}{c|}{768.7}           & 75.0          & 299.9          \\ \hline
LSC-CNN {\cite{ref52}}                 & PAMI 20                & {\ul 10.6}          & \multicolumn{1}{c|}{{\ul 31.8}}           & 34.9           & \multicolumn{1}{c|}{55.6}           & 601.9           & \multicolumn{1}{c|}{1172.2}          & 112.7         & 454.4          \\
CG-DRCN-CC-VGG16 {\cite{ref33}}        & PAMI 20                & 19.5          & \multicolumn{1}{c|}{58.7}           & 38.4           & \multicolumn{1}{c|}{62.7}           & 367.3           & \multicolumn{1}{c|}{837.5}           & 82.3          & 328.0          \\
CG-DRCN-CC-Res101 {\cite{ref33}}       & PAMI 20                & 14.0          & \multicolumn{1}{c|}{42.8}           & 35.0           & \multicolumn{1}{c|}{53.7}           & {\ul 314.7}     & \multicolumn{1}{c|}{{\ul 712.3}}     & 71.0          & 278.6          \\
KDMG {\cite{ref54}}                    & PAMI 20                & -             & \multicolumn{1}{c|}{-}              & -              & \multicolumn{1}{c|}{-}              & -               & \multicolumn{1}{c|}{-}               & 69.7          & 268.3          \\
NoisyCC {\cite{ref55}}                 & NeurIPS 20             & -             & \multicolumn{1}{c|}{-}              & -              & \multicolumn{1}{c|}{-}              & -               & \multicolumn{1}{c|}{-}               & 67.7          & {\ul 258.5}          \\
DM-Count {\cite{ref39}}                & NeurIPS 20             & -             & \multicolumn{1}{c|}{-}              & -              & \multicolumn{1}{c|}{-}              & -               & \multicolumn{1}{c|}{-}               & 68.4          & 283.3          \\ \hline
AutoScale {\cite{ref53}}               & IJCV 21                & 13.2          & \multicolumn{1}{c|}{\textbf{30.2}}     & {\ul 32.3}     & \multicolumn{1}{c|}{{\ul 52.8}}     & 425.6           & \multicolumn{1}{c|}{916.5}           & 85.6          & 356.1          \\
BM-Count {\cite{ref56}}                & IJCAI 21               & -             & \multicolumn{1}{c|}{-}              & -              & \multicolumn{1}{c|}{-}              & -               & \multicolumn{1}{c|}{-}               & 61.5          & 263.0          \\
URC {\cite{ref57}}                     & ICCV 21                & -             & \multicolumn{1}{c|}{-}              & -              & \multicolumn{1}{c|}{-}              & -               & \multicolumn{1}{c|}{-}               & 129.65        & 400.47         \\
TopoCount {\cite{ref45}}               & AAAI 21                & -             & \multicolumn{1}{c|}{-}              & -              & \multicolumn{1}{c|}{-}              & -               & \multicolumn{1}{c|}{-}               & 60.9          & 267.4          \\
GLoss {\cite{ref43}}                   & CVPR 21                & -             & \multicolumn{1}{c|}{-}              & -              & \multicolumn{1}{c|}{-}              & -               & \multicolumn{1}{c|}{-}               & {\ul 59.9} & 259.5          \\
D2C {\cite{ref44}}                    & TIP 21                 & 12.8             & \multicolumn{1}{c|}{38.6}        & 38.3              & \multicolumn{1}{c|}{58.7}        & 333.6           & \multicolumn{1}{c|}{751.6}           & 75.0          & 294.0          \\ \hline 
MAN {\cite{ref46}}                     & CVPR 22                & -             & \multicolumn{1}{c|}{-}              & -              & \multicolumn{1}{c|}{-}              & -               & \multicolumn{1}{c|}{-}               & \textbf{53.4} & \textbf{209.9}          \\
\hline \hline
Ours                    & -                      & 13.25 & \multicolumn{1}{c|}{\textbf{41.75}} & \textbf{30.32} & \multicolumn{1}{c|}{\textbf{51.97}} & \textbf{284.39} & \multicolumn{1}{c|}{\textbf{670.84}} & 63.47         & 262.64 \\ \hline 
\end{tabular}
\end{table*}

{\bf NWPU-Crowd } is currently the most challenging dataset in the field of crowd counting. It includes 5109 images and 2.13 million annotation points. The training set, val set and test set contain 3109, 500 and 1500 images, respectively. This dataset introduces 351 negative sample images, \textit{i.e.}, unoccupied scenes. The maximum number of targets for a single image up to $20033$, the average size is $2191 \times 3209$, and the maximum size reaches $4028 \times 19044$, which  requires high computing power. As can be seen from Table \textcolor{red}{\ref{Table_4}}, we rank first in the val set of the published methods. More critically, for the test set, we have surpassed this year's SOTA method by  3.28\% on the MSE metric, but are 7.3\% behind in MAE compared to it. Detailed, for different scene  $(S0 \sim S4)$   and luminance  $(L0 \sim L2)$  categories, we are 18.94\% and 7.5\% ahead of the  previou SOTA method in $S4$ and $L2$, respectively; in $S0$, we also reach the optimal level, outperforming the second place by 14.89\% .

\begin{table*}
\caption{\centering Comparison with state-of-the-art methods on the NWPU-CROWD dataset.  \textit{S0 $\sim$ S4} respectively indicates five categories according to the different number range: $0$, $(0, 100]$, $...$, $\geq 5000$. \textit{L0 $\sim$L2} respectively denotes three luminance levels: $[0, 0.25]$, $(0.25, 0.5]$, and $(0.5, 0.75]$. Limited by the paper length, only MAE are reported in the category-wise results. The best performance is indicated by {\bf bold} and the second best is {\ul underlined}.}  
\label{Table_4}
\centering
\scalebox{0.85}{
\begin{tabular}{|l|l|cc|ccccccccccccc|}
\hline
\multirow{3}{*}{Method} & \multirow{3}{*}{Venue} & \multicolumn{2}{c|}{Val set}         & \multicolumn{13}{c|}{Test set}                                                                                                                                                                                                                                  \\ \cline{3-17} 
                        &                        & \multicolumn{2}{c|}{Overall}         & \multicolumn{3}{c|}{Overall}                                          & \multicolumn{6}{c|}{Scene Level (MAE)}                                                                                 & \multicolumn{4}{c|}{Luminance (MAE)}                           \\ \cline{3-17} 
                        &                        & MAE              & MSE               & MAE           & MSE             & \multicolumn{1}{c|}{NAE}            & Avg.            & S0            & S1           & S2            & S3             & \multicolumn{1}{c|}{S4}              & Avg.          & L0             & L1            & L2            \\ \hline
MCNN {\cite{ref14}}                   & CVPR 16                & 218.5            & 700.6             & 232.5         & 714.6           & \multicolumn{1}{c|}{1.063}          & 1171.9          & 356.0         & 72.1         & 103.5         & 509.5          & \multicolumn{1}{c|}{4818.2}          & 220.9         & 472.9          & 230.1         & 181.6         \\ \hline
CSRNet {\cite{ref16}}                 & CVPR 18                & 104.8            & 433.4             & 121.3         & 387.8           & \multicolumn{1}{c|}{0.604}          & 522.7           & 176.0         & 35.8         & 59.8          & 285.8          & \multicolumn{1}{c|}{2055.8}          & 112           & 232.4          & 121.0         & 95.5          \\
SANet {\cite{ref47}}                  & ECCV 18                & -                & -                 & 190.6         & 491.4           & \multicolumn{1}{c|}{0.991}          & 716.3           & 432.0         & 65.0         & 104.2         & 385.1          & \multicolumn{1}{c|}{2595.4}          & 153.8         & 254.2          & 192.3         & 169.7         \\
PCC-Net {\cite{ref22}}                & T-CSVT 19              & 100.7            & 573.1             & 112.3         & 457.0           & \multicolumn{1}{c|}{0.251}          & 777.6           & 103.9         & 13.7         & 42.0          & 259.5          & \multicolumn{1}{c|}{3469.1}          & 111.0         & 251.3          & 111.0         & 82.6          \\
BL {\cite{ref37}}                     & ICCV 19                & 93.6             & 470.3             & 105.4         & 454.2           & \multicolumn{1}{c|}{0.203}          & 750.5           & 66.5          & 8.7          & 41.2          & 249.9          & \multicolumn{1}{c|}{3386.4}          & 115.8         & 293.4          & 102.7         & 68.0          \\
CAN {\cite{ref49}}                    & CVPR 19                & 93.5             & 489.9             & 106.3         & 386.5           & \multicolumn{1}{c|}{0.295}          & 612.2           & 82.6          & 14.7         & 46.6          & 269.7          & \multicolumn{1}{c|}{2647.0}          & 102.1         & 222.1          & 104.9         & 82.3          \\
SFCN {\cite{ref48}}                   & CVPR 19                & 95.4             & 608.3             & 105.4         & 424.1           & \multicolumn{1}{c|}{0.254}          & 712.7           & 54.2          & 14.8         & 44.4          & 249.6          & \multicolumn{1}{c|}{3200.5}          & 106.8         & 245.9          & 103.4         & 78.8          \\ \hline
DM-Count {\cite{ref39}}               & NeurIPS 20             & 70.5             & {\ul 357.6}       & 88.4          & 388.6           & \multicolumn{1}{c|}{\textbf{0.169}} & 498.0           & 146.7         & \textbf{7.6} & \textbf{31.2} & 228.7          & \multicolumn{1}{c|}{2075.8}          & 88.0          & 203.6          & 88.1          & 61.2          \\
NoisyCC {\cite{ref55}}                & NeurIPS 20             & -                & -                 & 96.9          & 534.2           & \multicolumn{1}{c|}{-}              & -               & -             & -            & -             & -              & \multicolumn{1}{c|}{-}               & -             & -              & -             & -             \\ \hline
BM-Count {\cite{ref56}}               & IJCAI 21               & -                & -                 & 83.4          & 358.4           & \multicolumn{1}{c|}{-}              & -               & -             & -            & -             & -              & \multicolumn{1}{c|}{-}               & -             & -              & -             & -             \\
TopoCount {\cite{ref45}}              & AAAI 21                & -                & -                 & 107.8         & 438.5           & \multicolumn{1}{c|}{-}              & -               & -             & -            & -             & -              & \multicolumn{1}{c|}{-}               & -             & -              & -             & -             \\
DKPNet {\cite{ref42}}                 & ICCV 21                & {\ul 61.8}       & 438.7             & \textbf{74.5} & 327.4           & \multicolumn{1}{c|}{-}              & -               & -             & -            & -             & -              & \multicolumn{1}{c|}{-}               & -             & -              & -             & -             \\
AutoScale {\cite{ref53}}              & IJCV 21                & 97.3             & 571.2             & 123.9         & 515.5           & \multicolumn{1}{c|}{-}              & 871.2           & {\ul 42.3}    & 18.8         & 46.1          & 301.7          & \multicolumn{1}{c|}{3947.0}          & 127.1         & 301.3          & 122.2         & 86.0          \\
D2C {\cite{ref44}}                    & TIP 21                 & -                & -                 & 85.5          & 361.5           & \multicolumn{1}{c|}{0.221}          & 539.9           & 52.4          & 10.8         & 36.2          & 212.2          & \multicolumn{1}{c|}{2387.8}          & {\ul 82.0}    & \textbf{177.0} & 83.9          & 68.2          \\
Gloss {\cite{ref43}}                  & CVPR 21                & -                & -                 & 79.3          & 346.1           & \multicolumn{1}{c|}{0.180}           & 508.5           & 92.4          & {\ul 8.2}    & 35.4          & \textbf{179.2} & \multicolumn{1}{c|}{2228.3}          & 85.6          & 216.6          & {\ul 78.6}    & {\ul 48.0}    \\ \hline
MAN {\cite{ref46}}                    & CVPR 22                & -                & -                 & {\ul 76.5}    & {\ul 323.0}     & \multicolumn{1}{c|}{{\ul 0.170}}     & {\ul 464.6}     & 43.3          & 8.5          & {\ul 35.3}    & {\ul 190.1}    & \multicolumn{1}{c|}{{\ul 2044.9}}    & \textbf{76.4} & {\ul 180.1}    & \textbf{77.1} & 49.4          \\ \hline \hline
Ours                    & -                      & \textbf{56.47} & \textbf{201.85} & 82.52         & \textbf{312.39} & \multicolumn{1}{c|}{0.300}           & \textbf{393.3} & \textbf{36.0} & 12.0         & 48.3          & 212.6          & \multicolumn{1}{c|}{\textbf{1657.5}} & 93.8          & 249.0          & 81.8          & \textbf{44.4} \\ \hline
\end{tabular}
}
\end{table*}

\subsection{ Ablation Study}

To explain the excellent performance of our  network in  the above benchmark datasets, we dissect in detail the contribution of its basic components and hyperparameters to the overall . In this section, we first observe the huge  gains brought by the SoftMax-Attention strategy to the network and the change in gains  when stacking different numbers of IIAO modules, then discuss the effectiveness of RCLoss and   $threshold$  , and finally explore the necessity of sliding windows with overlap  for RCLoss. To be fair, all the following experiments are conducted on the JHU-Crowd++ dataset (val set).

\begin{figure*}[!t]
\centering
\includegraphics[scale=0.49]{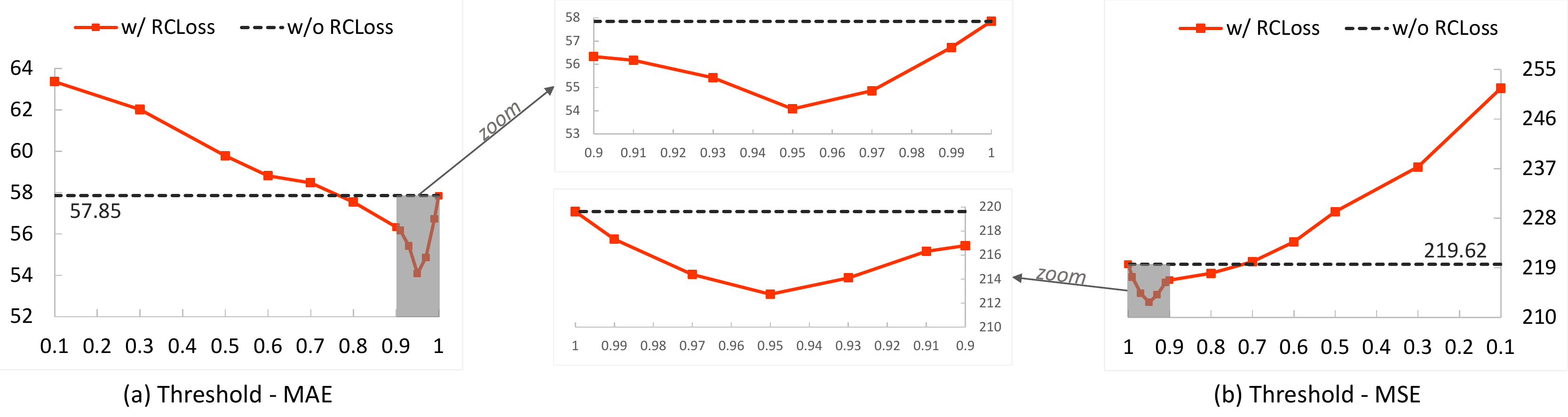}
\caption{  Effect of RCLoss and its $threshold$  on network prediction. The black dotted line indicates the experiment that replaces RCLoss with MSELoss, considered as  $benchmark$, which is equivalent to the case of $threshold = 1$ in RCLoss. The result is better with RCLoss only when $ 0.8 < threshold < 1$.}
\label{Figure_7}
\end{figure*}

\begin{table}[ht]
\caption{\centering The effect of different $F_{att}$  dimensionality reduction methods on network performance.}
\label{Table_5}
\centering
\scalebox{0.82}{
\begin{tabular}{|c|c|cc|cc|}
\hline
\rowcolor[HTML]{C0C0C0} 
\cellcolor[HTML]{C0C0C0}                        & \cellcolor[HTML]{C0C0C0}                         & \multicolumn{2}{c|}{\cellcolor[HTML]{C0C0C0}Loss Function} & \cellcolor[HTML]{C0C0C0}                      & \cellcolor[HTML]{C0C0C0}                      \\ \cline{3-4}
\rowcolor[HTML]{C0C0C0} 
\multirow{-2}{*}{\cellcolor[HTML]{C0C0C0}Group} & \multirow{-2}{*}{\cellcolor[HTML]{C0C0C0}Method} & \textit{MSELoss}                        & \textit{RCLoss}                        & \multirow{-2}{*}{\cellcolor[HTML]{C0C0C0}MAE $\downarrow$} & \multirow{-2}{*}{\cellcolor[HTML]{C0C0C0}MSE $\downarrow$} \\ \hline
-                                               & Baseline                                         & \Checkmark                          & \XSolidBrush                             & 79.62                                         & 266.50                                        \\ \cline{1-2}
1                                               & Dimensional averaging                            & \Checkmark                          & \XSolidBrush                             & 109.14                                        & 320.54                                        \\ \cline{1-2}
2                                               & Dimensional convolution                          & \Checkmark                          & \XSolidBrush                             & 65.39                                         & 241.60                                        \\ \cline{1-2}
3                                               & SoftMax-Attention                                & \Checkmark                          & \XSolidBrush                             & 57.85                                         & 219.62                                        \\ \hline
$\ell$                                               & SoftMax-Attention                                & \Checkmark                          & \Checkmark                             & 54.08                                         & 212.73                                        \\ \hline
\end{tabular}
}
\end{table}

\begin{figure}[!t]
\centering
\includegraphics[scale=0.47]{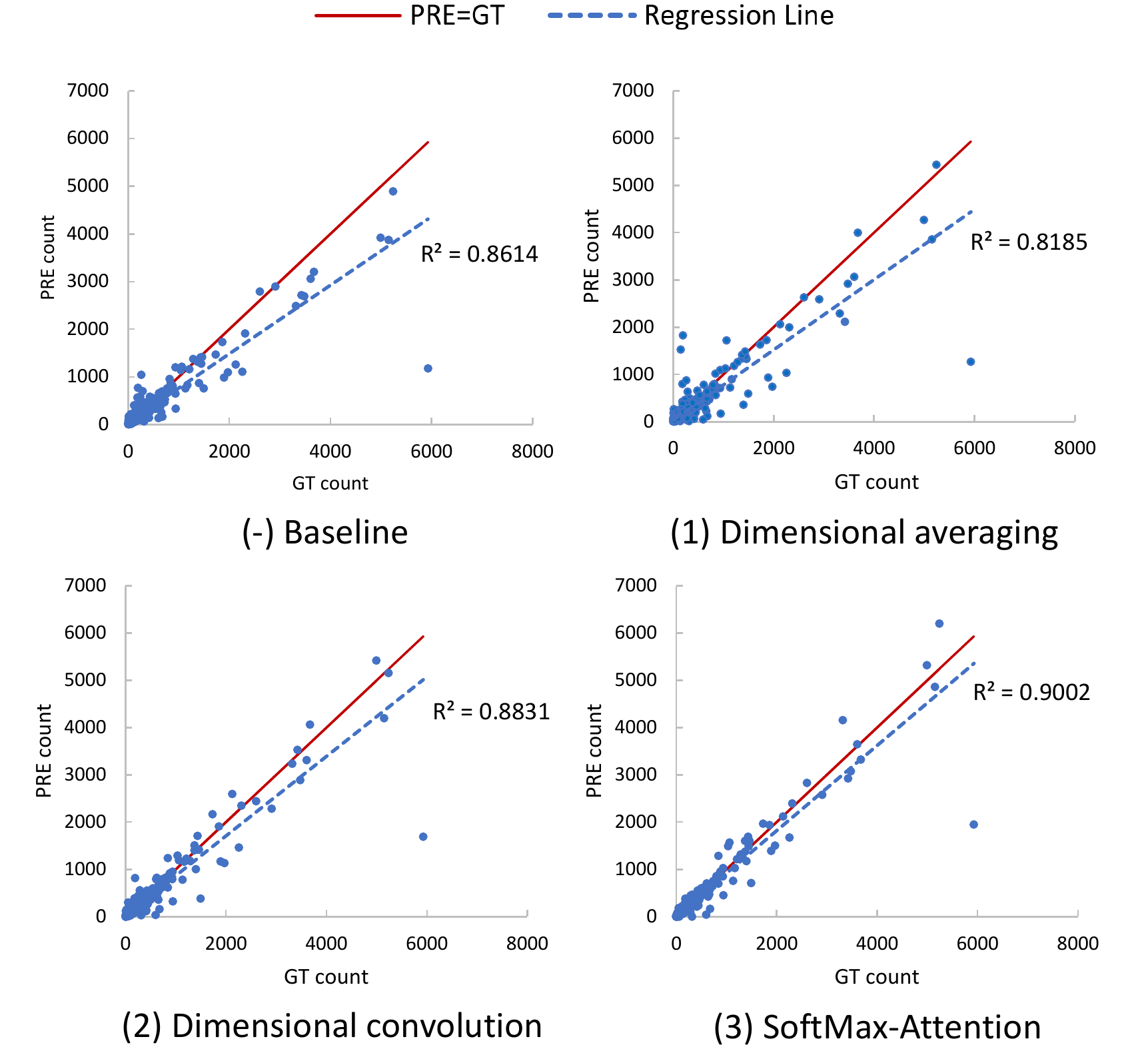}    
\caption{  Scatter diagram of predicted counts versus ground-truth counts on the JHU-Crowd++ dataset (val set). The blue dotted line is the regression result  of the scatter, and R-square ($R^2$) is its coefficient of determination; the reddish-brown line serves as an auxiliary and represents the ideal case with 100\% counting accuracy.}  
\label{Figure_6}
\end{figure}

{\bf Superiority of the SoftMax-Attention Strategy:} We promote  SoftMax-Attention and experimentally verify its advantages. The first variant is the \textit{baseline} that discards the Soft Block, which can also be expressed as: \textit{baseline= Primary Feature Extractor + Adaptive Scale Pyramid + Transitive Attention Unit} , so  the \textit{baseline} does not compute losses on $F_{att}$, but allows it to learn itself during gradient back-propagated. Next, three sets of control experiments are then set up according to the different methods of dimensionality reduction for $F_{att}$, in the following order:  
\begin{enumerate}
\item{Averaging $F_{att}$ by  channel direction.} 
\item{Setting  $1 \times 1$ convolution to reduce the channel to 1.}
\item{Using the SoftMax-Attention strategy.}
\end{enumerate}
To control variables consistently, all experiments up to now, including \textit{baseline}, only use  MSELoss. As shown in Table \textcolor{red}{\ref{Table_5}}, all indicators in the three control trials showed a decreasing trend. However, Group 1 is less effective than  \textit{baseline},   which is to be expected since it defies the will of the attention mechanism. Both Group 2 and Group 3 are ahead of \textit{baseline}, while the lower counting error of the latter confirms our hypothesis that there is indeed room for optimizing the loss calculation process , and our proposed IIAO module based on  SoftMax-Attention plays a key role. Not negligibly, Group $\ell$ optimized for $F_{wei}$ using RCLoss achieves better results, suggesting that RCLoss is the icing on the cake. To intuitively demonstrate the improvement achieved by SoftMax-Attention relative to the \textit{baseline}, we select typical samples from each dataset, see Figure \textcolor{red}{\ref{Figure_5}}. Further, to observe the overall prediction performance of each experimental group on  JHU-Crowd++ dataset(val set) , we plot \textit{PRE-GT} scatter diagram with regression line containing all samples in this dataset, and the auxiliary line $y = x$ represents the ideal case where the counting prediction accuracy is 100\%. The aggregated results are shown in Figure \textcolor{red}{\ref{Figure_6}}. Qualitatively, the tightness of the regression line to $y = x$ symbolizes the quality of prediction (positive correlation); quantitatively, the closer $R^2$ is to $1$, the smaller the overall error fluctuation.

\begin{table}[ht]
\caption{\centering Effect of IIAO module stacking number on network ontology and prediction performance.  Indicating arrows represent a positive trend.  }
\label{Table_6}
\centering
\scalebox{0.8}{
\begin{tabular}{|c|c|c|c|c|c|}
\rowcolor[HTML]{C0C0C0} 
\hline
Stacking number & GFLOPs $\uparrow$ & \begin{tabular}[c]{@{}c@{}}Param size $\downarrow$ \\ (MB)\end{tabular} & \begin{tabular}[c]{@{}c@{}}Inference time $\downarrow$ \\ (ms)\end{tabular} & MAE $\downarrow$  & MSE $\downarrow$    \\ \hline
1               & 61.46  & 19.68                                                      & 7.602                                                          & 58.46 & 224.60 \\ \hline
2               & 72.53  & 24.10                                                      & 9.177                                                          & 54.08 & 212.73 \\ \hline
3               & 83.60  & 28.53                                                      & 11.476                                                         & 54.52 & 211.43 \\ \hline
4               & 94.67  & 32.96                                                      & 13.093                                                         & 54.87 & 212.61 \\ \hline
\end{tabular}
}
\end{table}

{\bf Influence of stacking number of IIAO modules:} As aforementioned, IIAO can optimize attention and improve model performance, but the following experiments illustrate that more stacks may not be better. For a more comprehensive consideration, in addition to the necessary \textit{MAE} and \textit{MSE} , we also add the monitoring of three indicators \textit{Param size}, \textit{GFLOPs} and \textit{Inference time}, which are closely related to the practical value. Table \textcolor{red}{\ref{Table_6}} shows the performance of the network when stacking different numbers of IIAO modules: the \textit{MAE} and \textit{MSE } have converged when the number is $2$,  and the other metrics are within acceptable zone at this point. Moving on, excessive intervention leads to over-smoothing of features, while the redundant parameters put a great strain on the computer hardware. Note that to avoid the impact of inconsistent image sizes or other accidental factors on the inference process, we perform $1000$ consecutive forward inferences on a $3 \times 400 \times 400$ patch, and finally take the average time as the \textit{Inference time}.

{\bf Effectiveness of the RCLoss and $\bm {threshold}$:} Regional error-proneness varies by density distribution, while the proposed RCLoss can provide region-dependent loss penalties. To verify its validity, we replace RCLoss with MSELoss, consider as the \textit{benchmark}; immediately after, revert to RCLoss and adjust the value of $threshold$ several times as a multi-group control trial. Figure \textcolor{red}{\ref{Figure_7}} visualizes the performance fluctuations caused by RCLoss on the model. To expand on this, if the $threshold$ is  too small,  more regions are penalized and even adjacent regions are linked, making it difficult to highlight high-frequency error locations and thus converging slowly. Conversely, if too large,  the penalty condition becomes so harsh that the RCLoss will infinitely degenerate into MSELoss. Furthermore, the precision requirements for floating-point numbers become excessively stringent. From the curve, for the JHU-Crowd++ dataset (val set), it is optimal when using RCLoss and setting the $threshold$ to $0.95$. Notably, the  density level varies from one dataset to another, so we add experiments to the respective optimal values of $threshold$ under different datasets, refer to Table \textcolor{red}{\ref{Table_7}}.

\begin{table}[ht]
\caption{\centering $Threshold$ optimal values for different datasets. \label{Table_7}}
\centering
\begin{tabular}{|c|c|}
\rowcolor[HTML]{C0C0C0} 
\hline
Dataset                & $Threshold$ \\ \hline
SHA                    & 0.95      \\ \hline
SHB                    & 0.92      \\ \hline
UCF\_CC\_50            & 0.97      \\ \hline
UCF-QNRF               & 0.95      \\ \hline
JHU-Crowd++            & 0.95      \\ \hline
NWPU-Crowd             & 0.95      \\ \hline
\end{tabular}
\end{table}

{\bf Necessity of the Sliding Window in RCLoss:} The crowd density in the real scene is unevenly distributed, and there may be multiple error-prone regions in a single image, which results in the points with the top $N$ error values being cross-regional and losing spatial correlation. Hence, we divide and conquer the image, using a sliding window to provide more continuity candidates for the model, where the key element  is the setting of the sliding stride and the window size. For the $50 \times 50$ $F_{wei}$, we have three requirements for the selection of these two parameters.
\begin{enumerate}
\item{Padding = $0$}
\item{The sliding window traverses all positions in $E$}
\item{The maximum overlap length between  subwindows is $8$  }
\end{enumerate}

\begin{figure}[!t]
\centering
\includegraphics[scale=0.215]{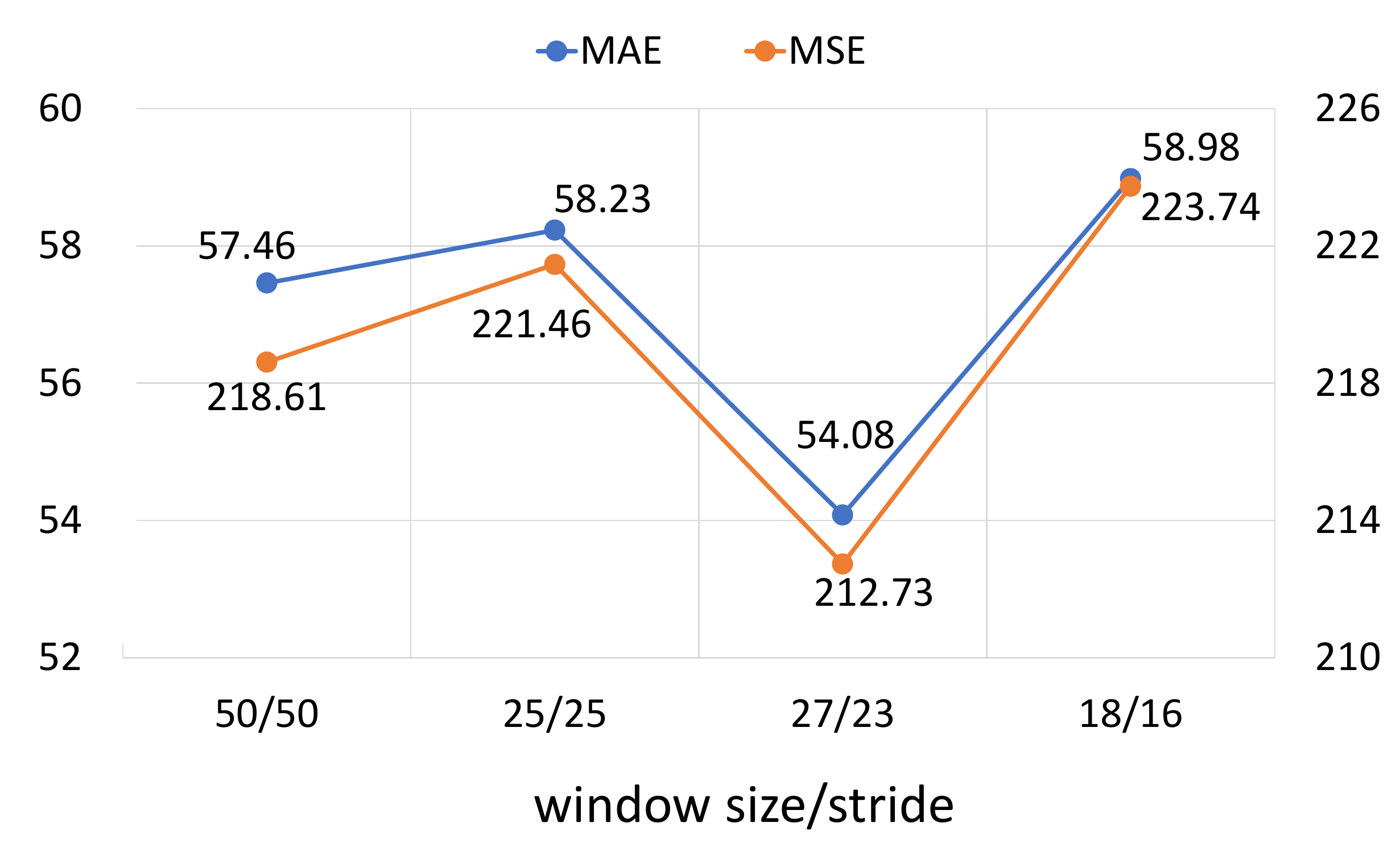}
\caption{ Effect of sliding stride and window size on network prediction.  } 
\label{Figure_8}
\end{figure}

With the above premise while keeping $threshold$ at $0.95$, we have experimented with several possible scenarios. The results are shown in Figure \textcolor{red}{\ref{Figure_8}}, which indicates that a size of $27$ and a stride of $23$ is the optimal combination. This is a window with an overlap length of $8$, and the known receptive field is also $8$; thus, the area mapped to the original image is $64 \times 64$, which is close to the size of a larger human head. While it is more likely that the larger head appears on the dividing line, in this case, this approach enables cross-image block analysis of features on the dividing line, thus avoiding possible targets from being corrupted.

\section{Conclusion }

In this paper, we dissect two potential loss calculation puzzles in traditional attention mechanisms: reliance on pseudo-labeling, and dimensionality reduction operations obfuscates attention parameter seeking optimization. Therefore, we innovatively propose the IIAO module based on SoftMax-Attention strategy, which transforms high-dimensional attention map into a  one-dimensional feature map in the mathematical sense for loss calculation midway through the network, while automatically providing adaptive multi-scale fusion for the feature pyramid module and relieving the pressure caused by the variable scale. We also propose RCLoss focusing on continuous error-prone regions to assist the IIAO modules in uncovering high-value regions, thus accelerating model convergence. Extensive experiments on several benchmark datasets show that our approach outperforms the previous SOTA methods.


\end{document}